\documentclass[10pt,twocolumn,letterpaper]{article}

\usepackage{cvpr}
\usepackage{times}
\usepackage{graphicx}
\usepackage{epsfig}
\usepackage{amsmath}
\usepackage{amssymb}
\usepackage{booktabs} % nice tables
\usepackage[table, dvipsnames]{xcolor}
\usepackage{paralist}
\usepackage{comment}
\usepackage[breaklinks=true,bookmarks=false, colorlinks=true]{hyperref}

\newcommand{\specialcell}[2][b]{%
  \begin{tabular}[#1]{@{}c@{}}#2\end{tabular}}

\cvprfinalcopy % *** Uncomment this line for the final submission

\begin{document}

%%%%%%%%% TITLE
\title{Object Tracking by Reconstruction with View-Specific Discriminative Correlation Filters}

\author{U\u{g}ur Kart$^1$, Alan Luke{\v{z}}i{\v{c}}$^2$, Matej Kristan$^2$, Joni-Kristian K\"am\"ar\"ainen$^1$ and Ji\v{r}\'{i} Matas$^3$\\
$^1$Laboratory of Signal Processing, Tampere University of Technology, Finland\\
$^2$Faculty of Computer and Information Science, University of Ljubljana, Slovenia \\
$^3$Faculty of Electrical Engineering, Czech Technical University in Prague, Czech Republic \\
{\tt\small \{ugur.kart, joni.kamarainen\}@tut.fi} \\
{\tt\small \{alan.lukezic, matej.kristan\}@fri.uni-lj.si} \\
{\tt\small matas@cmp.felk.cvut.cz}
}

%\author{\IEEEauthorblockN{\IEEEauthorrefmark{1},
%\IEEEauthorrefmark{2}, \IEEEauthorrefmark{2} and
%\IEEEauthorrefmark{4}}
%\IEEEauthorblockA{Department of Whatever,
%Whichever University\\
%Wherever\\
%Email: \IEEEauthorrefmark{1}author.one@add.on.net,
%\IEEEauthorrefmark{2}author.two@add.on.net,
%\IEEEauthorrefmark{3}author.three@add.on.net,
%}}

\maketitle
%\thispagestyle{empty}

%%%%%%%%% ABSTRACT
\begin{abstract}
Standard RGB-D trackers treat the target as an inherently 2D structure, 
which makes modelling  appearance changes related even to simple out-of-plane
rotation highly challenging. 
%A major limitation of the standard RGB-D trackers is that the target is inherently considered as a 2D structure, which makes dealing with appearance changes related even to a simple out-of-plane rotation highly challenging. 
We address this limitation by proposing a novel long-term RGB-D tracker -- Object Tracking by Reconstruction (OTR). The tracker 
performs online 3D target reconstruction to  facilitate robust learning
of a set of view-specific discriminative correlation filters (DCFs). 
The 3D reconstruction supports two performance-enhancing features:
(i) generation of accurate spatial support  for constrained DCF learning from its 2D projection and 
(ii) point-cloud based estimation of 3D pose change for selection and storage of
view-specific DCFs which are used to robustly localize the target after out-of-view rotation or heavy occlusion. 
%The proposed reconstruction based approach tackles out-of-plane rotations which are inherently challenging for the existing trackers. 
Extensive evaluation of OTR on the challenging Princeton RGB-D  tracking  and
STC Benchmarks shows it outperforms the \textit{state-of-the-art} by a large margin.

\end{abstract}

%%%%%%%%% BODY TEXT
\section{Introduction}

Visual object tracking (VOT) is one of the fundamental problems in computer vision and has  many applications~\cite{hu_surveillance,Chaumette_robot}.
The field has progressed rapidly, fueled by the  availability of large and diverse datasets~\cite{OTB,smeulders_pami_2014} 
and the annual VOT challenge~\cite{Vot2016,Vot2017}. 
Until recently, tracking research has focused on RGB videos, largely neglecting RGB-D (rgb+depth) tracking as 
obtaining a reliable depth map at video frame rates has not been possible without expensive hardware.
In the last few years, depth sensors have become widely accessible, which  has lead to a significant 
increase of RGB-D tracking related work~\cite{dskcf_bmvc,DLST,ca3dms}. Depth provides important cues for tracking since it simplifies reasoning about
occlusions and facilitates depth-based object segmentation. Progress in RGB-D tracking has been further boosted by the emergence of
standard datasets and evaluation protocols~\cite{princetonrgbd,STC}.
\begin{figure}[!t]
\centering
\includegraphics[width=1\linewidth]{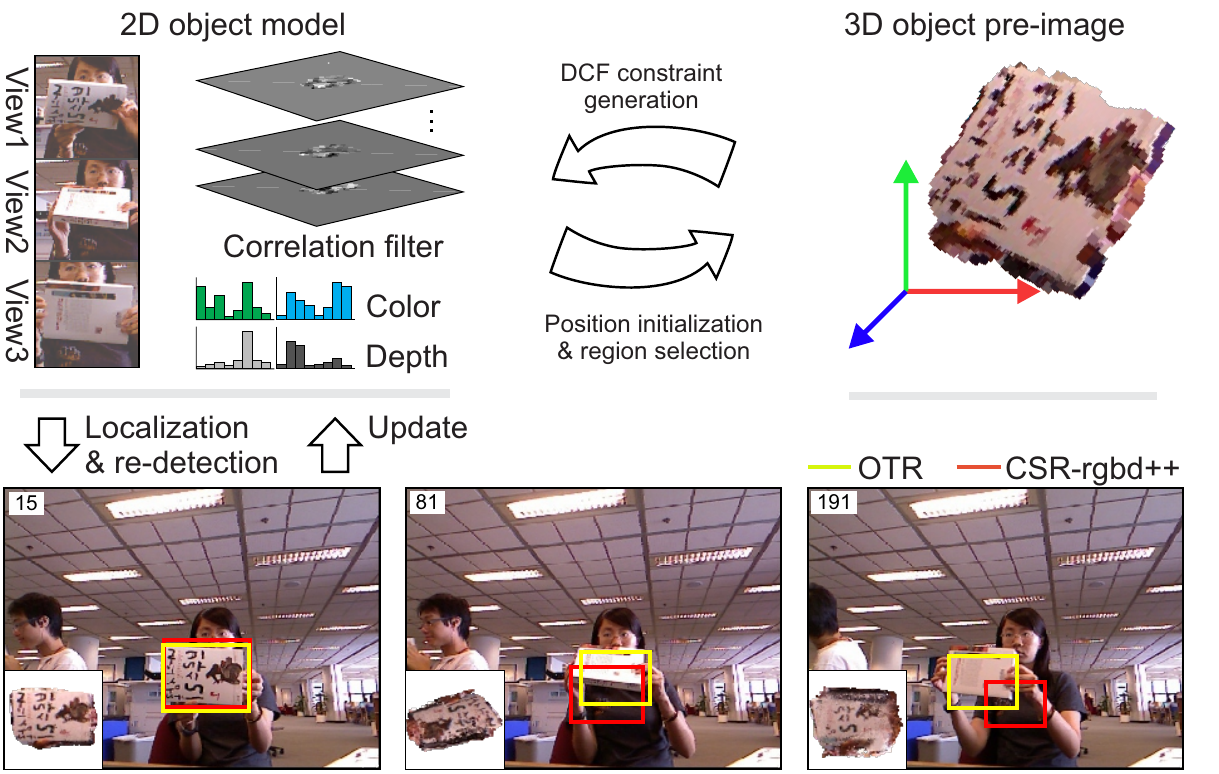}
\caption{The Object Tracking by Reconstruction (OTR)  object model
consists of a set of 2D view-specific DCFs
and of an approximate 3D object reconstruction (a 3D pre-image). The DCFs robustly localize the target and identify pixel regions for updating the 3D pre-image. In turn, the pre-image is used to constrain and guide appearance learning of the 2D filters. 
The OTR thus copes well with out-of-view rotation with a significant aspect change, while a state-of-the-art tracker CSR-rgbd++ drifts and fails.
%\cmnt{[AL] Constant interaction between 2D object model (a DCF, a multi-view DCF, color and depth target model) and 3D object pre-image provides better tracking results in a case of out-of-plane rotation than a DCF-based tracker without 3D model. An improvement in tracking performance due to the 2D-3D interation is demonstrated in the example below.}
}
\label{fig:intro_figure}
\end{figure}

In RGB-D tracking, direct extensions of RGB methods by adding the D-channel as an additional input dimension 
have achieved considerable success.
In particular, discriminative correlation filter (DCF) based methods have shown excellent performance
on the Princeton RGB-D tracking benchmark~\cite{princetonrgbd}, confirming the reputation gained on RGB benchmarks~\cite{Vot2016,Vot2017,Kart_ECCVW, DMDCF,dskcf_bmvc,DLST}.
Furthermore, DCFs are efficient in both learning of the visual target appearance model and in target localization, which are both implemented by FFT, running in near real time on a standard CPU. 

% 1 Critique/limits of the 2D representation in tracking, non DCF specific
% . Most RGB trackers model the object in 2D. 3D effects
%   -- appearance change due to pose change (rotation) has to be handle as apperance
%   -- example in Fig 1
%   -- aspect, outline changes difficult to handle, are unpredictable.
%   -- no info about 3D behaviour of the object, the pose is reported in 2D, as a corners of the boudnding box or per-pixel 
%      segmenation
% 2. We explooit the opportunity with D to build a simple and yet powerful representation, in term of surfels, proved
%   useful in the context of slam
%   -- we exploit ICP like process to build the model
%   -- to go for view-specific DCF
%, 3. advantages of 3D  handling visibility of only a part of the object, self-occlusion ....
%   -- model out-of-plane rotation
%   However, we can do 3D matching only for small motions 
%  4. View-specific DCFs give the redection, and thus long-term capability, regarless of the pose object

% 1
A major limitation of the standard RGB and RGB-D trackers, regardless of the actual method (e.g. DCF~\cite{Bolme-2010-cvpr}, Siamese deep nets~\cite{bertinetto2016fully}, Mean shift~\cite{meanshift}, Lucas Kanade~\cite{Lukas_Kanade}), is that they treat the tracked 3D object as a 2D structure.
Thus even a simple rotation of a rigid 3D object is interpreted as potentially significant appearance change in 2D that is conceptually indistinguishable from  partial occlusion, tracker drift, blurring and ambient light changes. 
 
Consider a narrow object, e.g., a book, with its front cover facing the camera, that rotates sideways and ends with its back side facing the camera (Figure~\ref{fig:intro_figure}). 
From the perspective of a standard RGB tacker, the object has deformed and the appearance has completely changed. Since most of the standard trackers cannot detect (do not model) aspect changes, the target bounding box and the
appearance model contain mostly pixels belonging to the background when the narrow side of the book is facing the camera. Furthermore, the model update is carried out by implicit or explicit temporal averaging of the tracked views.
Consequently, the appearance observed in the earlier frames is lost after a certain time period, limiting re-detection capability in situation when 
the target is completely occluded, but later re-appears, since its appearance no longer matches the last observed view.
The above-mentioned problems are almost trivial to solve if a photometric 3D model of the target object
is available.
%\cmnt{[UK] Should we change this part ? The ablation studies don't show any change in Occlusion category since majority of occlusion sequences in PTB are with non-rigid objects where 3D model doesn't work well. Reviewers might use this against us. Our formulation is very useful in long sequences where a rigid object goes under 360 degrees rotation and gets back to its initial pose so perhaps we can mention that.}

We exploit the opportunity of using the depth component in RGB-D signal to build a simple, yet powerful 3D object representation based on the surface splat model
(i.e., the object surface is approximated by a collection of 3D points with color, radius and the normal --  \textit{surfels}). 
This model has been proven very powerful in the context of SLAM~\cite{ruenz2017icra}. The 3D model is aligned and updated to the current 2D target appearance during tracking by an ICP-based matching mechanism~\cite{ruenz2017icra}
-- thus a \textit{pre-image} of the 2D target projection is maintained during tracking. The 3D object pre-image significantly simplifies detection and handling of (self-)occlusion, out-of-plane rotation (view changes) and aspect changes. 

The ICP-based 3D pre-image construction~\cite{ruenz2017icra} requires accurate identification of the object pixels in the current frame prior to matching, and it copes with only small motions due to a limited convergence range. 
A method from a high-performance RGB-D DCF tracker~\cite{Kart_ECCVW} is thus used to robustly estimate potentially large motions and to identify object pixels for the pre-image construction. The DCF learning is improved by generating appearance constraints from the pre-image. %and occlusion detection is improved by considering the outputs of both DCF and the pre-image.\cmnt{[UK]We don't use pre-image in occlusion detection/recovery. It's CSR segmentation}
Object appearance changes resulting from out-of-view rotation are detected 
by observing the pre-image 3D motion and a set of view-specific DCFs is generated. 
These 2D models are used during tracking for improved localization accuracy as well as for target re-detection using the recent efficient formulation of the DCF-based detectors~\cite{lukezicFCLT}. The resulting tracker thus exhibits a long-term capability, even if the target re-appears in a pose different from the one observed before the occlusion.

\paragraph{Contributions}
The main contribution of the paper is a new long-term RGB-D tracker, called OTR --  Object Tracking by Reconstruction,  that constructs a 3D model with view-specific DCFs attached.
The DCF-coupled estimation of the object pre-image and its use in DCF model learning for robust localization has not been proposed before. The OTR tracker achieves the \textit{state-of-the-art}, outperforming prior trackers by a large margin on two standard, challenging RGB-D tracking benchmarks.
An ablation study confirms the importance of view-specific DCF appearance learning that is tightly connected to the 3D reconstruction. We plan to make the reference implementation of OTR publicly available.

\section{Related Work}

\paragraph{RGB Tracking} 
Of the many approaches proposed in the literature, DCF-based methods have demonstrated excellent performance -- efficiency trade-off in recent tracking challenges~\cite{Vot2014,Vot2016,Vot2017}. 
Initially proposed by Bolme \etal~\cite{Bolme-2010-cvpr}, DCF-based tracking captured the attention of the vision community due to its simplicity and mathematical elegance.~Improvements of the original method include multi-channel formulation of correlation filters~\cite{danelljan2014adaptive,galoogahi_multi_channel_correlation}, filter learning using kernels exploiting properties of circular correlation~\cite{Henriques_KCF} and scale estimation with multiple one-dimensional filters~\cite{danelljan_dsst_pami}. 
Following these developments, Galoogahi~\etal~\cite{Galoogahi-2015-cvpr}
tackled the boundary problems that stem from the nature of circular correlation by proposing a filter learning method where a filter with size smaller than the training example is adopted.
Lukezic~\etal~\cite{csr} further improved this idea by formulating the filter learning process using a graph cut based segmentation mask as a constraint.

\vspace{-\medskipamount}
\paragraph{RGB-D Tracking} 
The most extensive RGB-D object tracking benchmark has been proposed by Song~\etal~\cite{princetonrgbd} (Princeton Tracking Benchmark). The benchmark includes a dataset, evaluation protocol and a set of baseline RGB-D trackers. 
Several RGB-D trackers have been proposed since. Meshgi~\etal~\cite{MESHGI_OAPF} used an occlusion-aware particle filter framework. 
A similar approach was proposed by Bibi~\etal~\cite{Bibi3D} but using optical flow to improve localization accuracy.
As an early adopter of DCF based RGB-D trackers, Hannuna~\etal~\cite{Hannuna2016} used depth as a clue to detect occlusions while tracking is achieved by KCF~\cite{Henriques_KCF}. 
An~\etal~\cite{DLST} performed a depth based segmentation along with a KCF tracker. Kart~\etal~\cite{DMDCF} proposed a purely depth based segmentation to train a constrained DCF similarly to CSR-DCF~\cite{csr} and later extended their work to include color in segmentation~\cite{Kart_ECCVW}. 
Liu~\etal~\cite{ca3dms} proposed a context-aware 3-D mean-shift tracker with occlusion handling. At the time of writing this paper~\cite{ca3dms} is ranked first at Princeton Tracking Benchmark.
Xiao~\etal~\cite{STC} recently proposed a new RGB-D tracking dataset (STC) and an RGB-D tracker by adopting an adaptive range-invariant target model.

\vspace{-\medskipamount}
\paragraph{3D Tracking} 
Klein~\etal~\cite{PTAM} proposed a camera pose tracking algorithm for small workspaces which works on low-power devices. The approach is based on tracking keypoints across the RGB frames and bundle adjustment for joint estimation of the 3D map and camera pose. 
Newcombe~\etal~\cite{Kinect_Fusion} proposed an iterative closest point (ICP) based algorithm for depth sequences for dense mapping of indoor scenes.~In a similar fashion, Wheelan~\etal~\cite{Elastic_Fusion} used surfel-based maps and jointly optimized color and geometric costs in a dense simultaneous localization and mapping (SLAM) framework. 
All three methods are limited to static scenes and are inappropriate for object tracking. 
This limitation was addressed by R{\"u}nz~\etal~\cite{ruenz2017icra}, who extended~\cite{Elastic_Fusion} by adding the capability of segmenting the scene into multiple objects. They use a motion consistency and semantic information to separate the object from the background. This limits the method to large, slow moving objects.

%As an attempt to solve out-of-plane rotation problem, 
Lebeda~\etal~\cite{Lebeda14} combined structure from motion, SLAM and 2D tracking to cope with 3D object rotation.
Their approach reconstructs the target by tracking keypoints and line features, however, it cannot cope with poorly-textured targets and low-resolution images.

\section{Object tracking by 3D reconstruction}\label{sec:conclusion}

%\textcolor{red}{A tracker needs to model the target object appearance to localize the object in the next frame and to re-detect the object in the case of tracking failure. In the proposed framework \ldots}

In OTR, object appearance is modeled at two levels of abstraction which
enables per-frame target localization and re-detection in the case of tracking failure.
The appearance level used for localizing the target in the image is modelled by a a set of view-specific discriminative correlation filters, i.e., a DCF $\mathbf{h}_t$ that models the current object appearance, and a set of snapshots $\{ \mathbf{h}^{(s)} \}_{s=1}^{S}$ modelling the object from previously observed views. In addition to the filters, the object color and depth statistics are modelled by separate color and depth histograms for 
the foreground and the background.

The second level of object abstraction is a model of the object pre-image $\Theta_t = \{ \mathbf{P}_t, \mathbf{R}_t, \mathbf{T}_t \}$, where $\mathbf{P}_t$ is the surfel-based object 3D model specified in the object-centered coordinate system and $\{\mathbf{R}_t, \mathbf{T}_t\}$ are the rotation and the translation of the 3D model into the current object position.

The two models interact during tracking for improved DCF training and 3D pose 
change detection (e.g., rotations). We describe the DCF framework used 
by the OTR tracker in Section~\ref{sec:csrdcf}, the multi-view DCFs with
the pre-image model is detailed in Section~\ref{sec:mvdcf}, Section~\ref{sec:target_redetection} details target loss recovery and Section~\ref{sec:tracker-summary} summarizes the full per-frame tracking iteration.
 
\subsection{Constrained DCF}\label{sec:csrdcf}

The core DCF tracker in the OTR framework is the recently proposed constrained discriminative correlation filter CSR-DCF~\cite{LukezicIJCV}, which is briefly outlined here. Given a search region of size $W \times H$ a set of $N_d$ feature channels $\mathbf{f} = \{ \mathbf{f}_d \}_{d=1}^{N_d}$, where $\mathbf{f}_d\in \mathcal{R}^{W\times H}$, are extracted. A set of $N_d$
correlation filters $\mathbf{h} = \{ \mathbf{h}_d \}_{d=1}^{N_d}$, where $\mathbf{h}_d\in \mathcal{R}^{W\times H}$, are correlated with the extracted features and the object position is estimated as the location of the maximum of the weighted correlation responses
\begin{equation}\label{eq:cf_localization}
	\mathbf{r}= \sum\nolimits_{d=1}^{N_d} w_d (\mathbf{f}_d \star \mathbf{h}_d),
\end{equation}
where $\star$ represents circular correlation, which is efficiently implemented by a Fast Fourier Transform with $\{ w_d \}_{d=1}^{N_d}$ being the channel weights. The target scale can be efficiently estimated by another correlation filter trained over the scale-space~\cite{danelljan_dsst_pami}. %\textcolor{red}{The estimated scale factor is used to resize the filters ($W\times H \rightarrow W' \times H'$).} {[MK] H',W' are not defined, nor is it true that the filters are resized -- the filters are never resized. Scale adaptation is technically more complicated and I do not see any reason to elaborate it here since it's not our contribution.
%}

Filter learning is formulated in CSR-DCF as a constrained optimization that minimizes a regression loss
\begin{equation}\label{eq:cf_cost_f_indep}
	\mathbf{\varepsilon}(\mathbf{h}) = \sum_{d=1}^{N_c} \| \mathbf{f}_d \star \mathbf{h}_d - \mathbf{g} \|^2 + \lambda \| \mathbf{h}_d \|^2 ~;~ \mathbf{h}_d \equiv \mathbf{m} \odot \mathbf{h}_d,
\end{equation}
where $\mathbf{g}$ is a desired output and $\mathbf{m}$ is a binary mask $\mathbf{m}\in \{0,1\}^{W \times H}$ that approximately separates the target from the background. The mask thus acts as a constraint on the filter support, which allows learning a filter from a larger training region as well as coping with targets that are poorly approximated by an axis-aligned bounding box. CSR-DCF applies a color histogram-based segmentation for mask generation, which is not robust to visually similar backgrounds and illumination change. We propose generating the mask from the RGB-D input and the estimated pre-image in Section~\ref{sec:filter_constraint}. 

Minimization of (\ref{eq:cf_cost_f_indep}) is achieved by an efficient ADMM scheme~\cite{admm_boyd2011}. Since the support of the learned filter is constrained to be smaller than the learning region, the maximum response on the training region reflects the reliability of the learned filter~\cite{csr}. 
These values are used as per-channel weights $w_d$ in (\ref{eq:cf_localization}) for improved target localization (we refer the reader to ~\cite{LukezicIJCV} for more details).

\subsection{A multi-view object model} \label{sec:mvdcf}

At each frame, the current filter $\mathbf{h}_t$ is correlated within a search region centered on the target position predicted from the previous frame following (\ref{eq:cf_localization}). 
To improve localization during target 3D motion, we introduce a "memory" which is implemented by storing captured snapshots $\{ \mathbf{h}^{(s)} \}_{s=1}^{S}$ from different 3D view-points (i.e.,~a set of view-specific DCFs). 
At every $N_\mathrm{R}$-th frame, all view-specific DCFs are evaluated, and the location of the maximum of the correlation response is used as the new target hypothesis $\mathbf{x}_t$.
If the maximum correlation occurs in the set of snapshots, the current filter is replaced by the corresponding snapshot filter.
%The new target position hypothesis $\mathbf{x}_t$ is estimated as the location of the maximum of the correlation response.
Target presence is determined at this location by the test described in Section~\ref{sec:target_presence_test}. In the case the test determines target is lost, the tracker enters a re-detection stage described in Section~\ref{sec:target_redetection}.

If the target is determined to be present, the current filter  $\mathbf{h}_t$
%is replaced by the snapshot that yields a better response (if it exists) and the %resulting filter $\mathbf{h}_t$
is updated by a weighted running average
\begin{equation} \label{eq:filter-update}
\mathbf{h}_{t+1} = (1-\eta)\mathbf{h}_{t} + \eta \tilde{\mathbf{h}}_{t},
\end{equation}
where $\tilde{\mathbf{h}}_{t}$ is a new filter estimated by the
constrained filter learning in Section~\ref{sec:csrdcf} at the estimated position $\mathbf{x}_t$ and $\eta$ is the update factor.

In addition to updating the current filter, the object color and depth histograms are updated as in~\cite{Kart_ECCVW}, the object pre-image is updated as described in Section~\ref{sec:preimage_update} and the set of view-specific DCFs $\{ \mathbf{h}^{(s)} \}_{s=1}^{S}$ is updated following Section~\ref{sec:mv_snapshots}.
 
\subsubsection{Object pre-image-based filter constraint}  \label{sec:filter_constraint} 

The binary mask $\mathbf{m}$ used in the constrained learning in (\ref{eq:cf_cost_f_indep}) is computed at the current target position at filter learning stage. 
In the absence of other inputs, the mask is estimated by a recent segmentation approach from~\cite{Kart_ECCVW}. 
This approach uses an MRF segmentation model from CSR-DCF~\cite{LukezicIJCV} within the filter learning region and estimates per-pixel unary potentials by color and depth (foreground/background) histograms backprojection in the RGB-D image.
   
\begin{figure}[!t]
\centering
\includegraphics[width=1\linewidth]{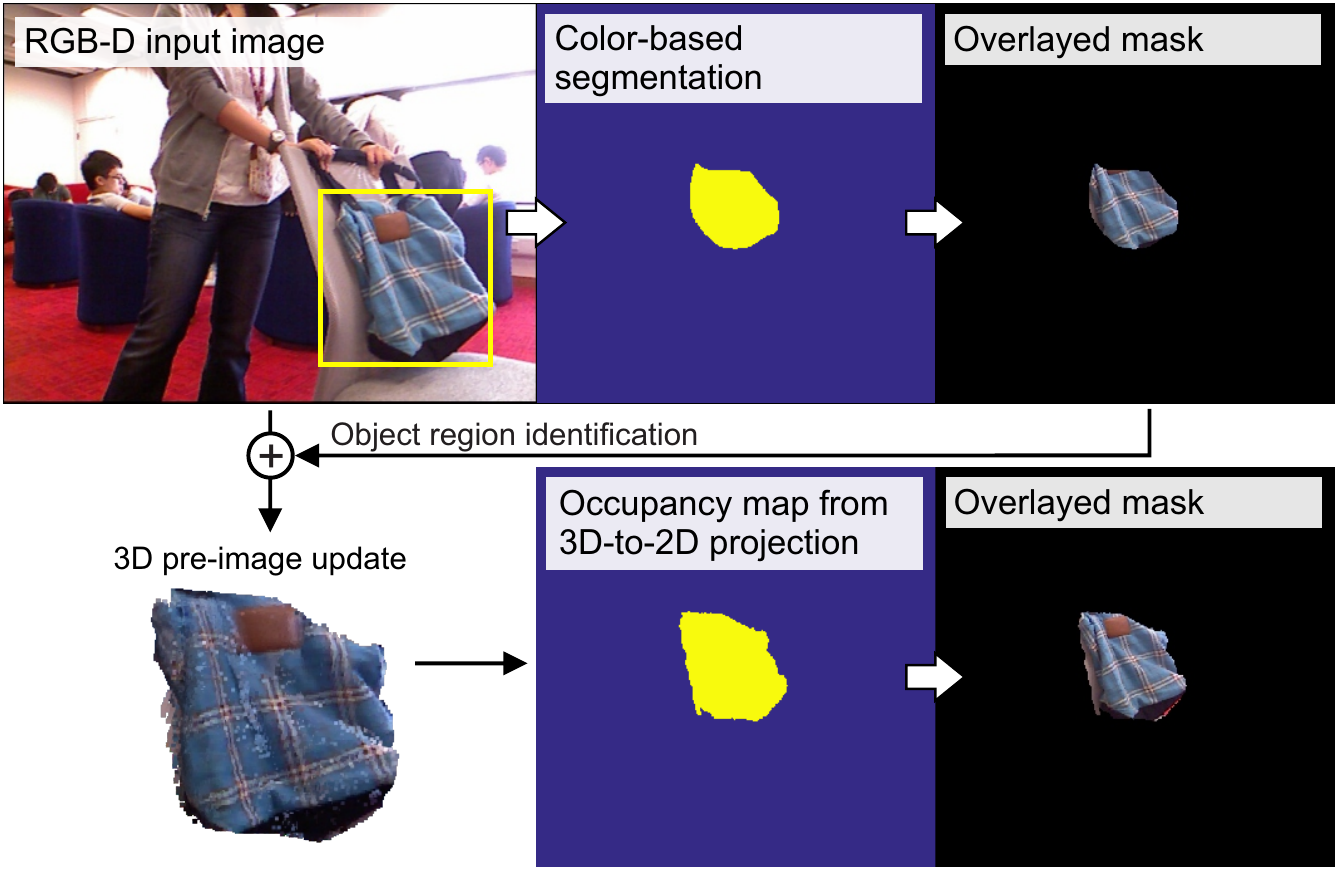}
\caption{A 2D DCF localizes the target (top-left), the target color+depth pixels are approximately segmented (top-right) and used to update the 3D pre-image (bottom-left). The pre-image is projected to 2D generating an occupancy map (bottom-right). The resulting mask better delineates the object, which improves the constrained DCF learning.
%The 3D model provides better segmentation mask than color and depth based segmentation. Binary segmentation mask is used as a constraint in filter learning.\cmnt{[UK] It should be "binary mask" since we don't use segmentation in projection}
}
\label{fig:preimage-mask}
\end{figure}

However, the pre-image $\Theta_t$ can be used to better outline the object in the filter training region, leading to a more accurately learned filter. 
Thus, at DCF training stage, the pre-image is generated by fitting the object 3D model $\mathbf{P}_t$ onto the current object appearance (Section~\ref{sec:preimage_update}). 
If the fit is successful, the segmentation mask used in filter learning (\ref{eq:cf_cost_f_indep}) is replaced by a new mask generated as follows. 
The 3D model $\mathbf{P}_t$ is projected into the 2D filter training region. Pixels in the region corresponding to the visible 3D points are set to one, while others to zero, thus forming a binary object occupancy map. 
The map is dilated to remove holes in the object mask and only the largest connected component is retained, while others are set to zero to reduce the effect of potential reconstruction errors in the 3D model. 
An example of the 2D mask construction from the 3D object pre-image is demonstrated in Figure~\ref{fig:preimage-mask}.

\subsubsection{Object pre-image update} \label{sec:preimage_update}

The object pre-image $\Theta_t$ is updated from the object position estimated by the multi-view DCF (Section~\ref{sec:mvdcf}). 
Pixels corresponding to the target are identified by the color+depth segmentation mask from Section~\ref{sec:filter_constraint}. 
The patch is extracted from the RGB-D image and used to update the object 3D model $\mathbf{P}_t$. 
The 3D model $\mathbf{P}_t$ is first translated to the 3D position determined by the target location from the multi-view DCF. 
The ICP-based fusion from~\cite{ruenz2017icra}, that uses color and depth, is then applied to fine align the 3D model with the patch and update it by adding and merging the corresponding surfels (for details we refer to~\cite{ruenz2017icra}). 
The updated model is only retained if the ICP alignment error is reasonably low (i.e., below a threshold $\tau_\mathrm{ICP}$), otherwise the update is discarded.

\subsubsection{A multi-view DCF update}\label{sec:mv_snapshots}

Continuous updates may lead to gradual drift and failure whenever the target object undergoes a significant appearance change. Recovery from such situations essentially depends on the diversity of the target views captured by the snapshots $\{ \mathbf{h}^{(s)} \}_{s=1}^{S}$ and their quality (e.g., snapshots should not be contaminated by the background). The following conservative update mechanism that maximizes snapshot diversity and minimizes contamination is applied. 

\begin{figure}[t]
\centering
\includegraphics[width=1\linewidth]{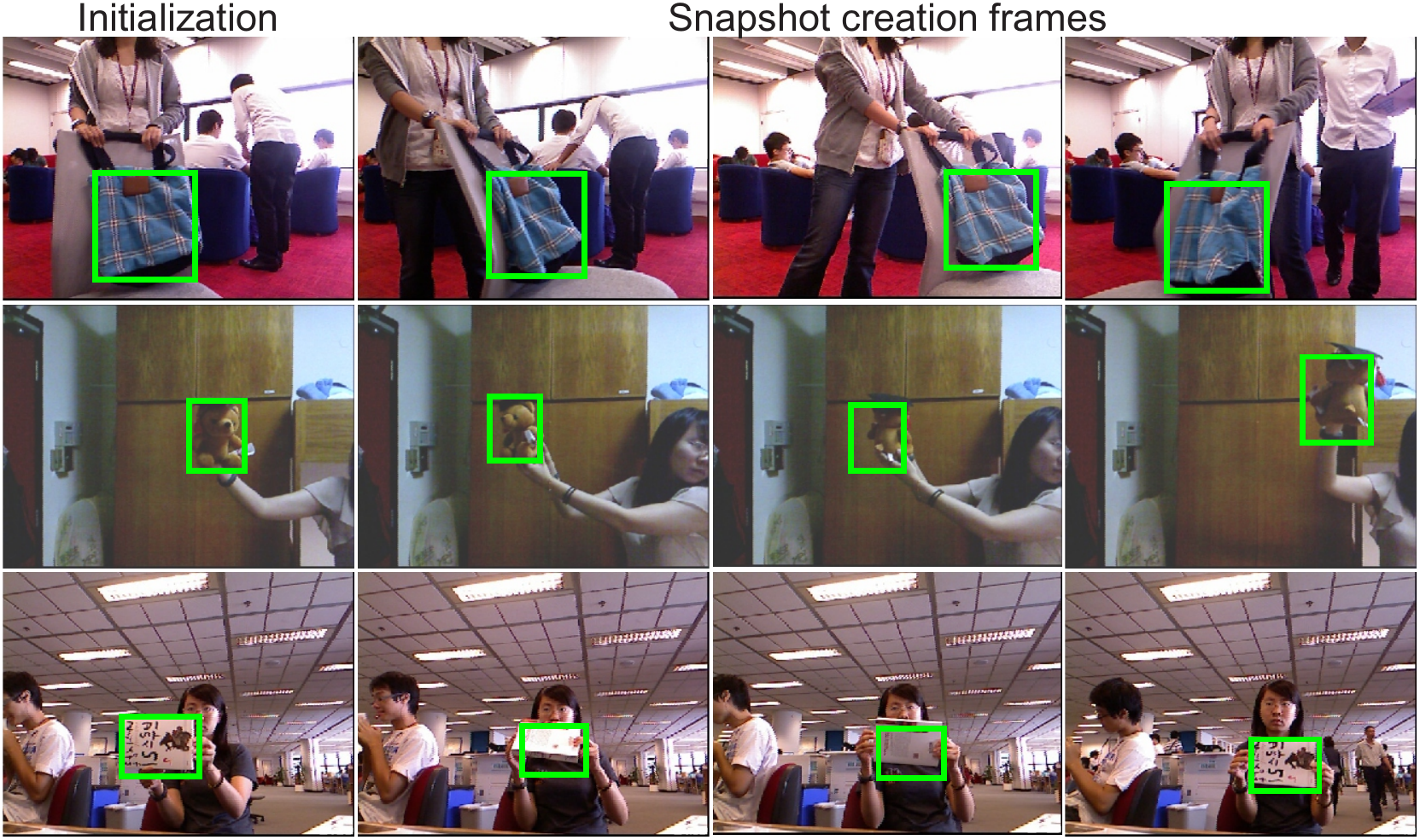}
\caption{Examples of view-specific DCFs creation. The tracker was initialized on the images in the left-most column, while the remaining images represent frames in which a new view was detected and stored in the set of view-specific DCFs.}
\label{fig:snapshots}
\end{figure}

The current filter is considered for addition to the snapshots only if the target passed the presence test (Section~\ref{sec:target_presence_test}) and the object pre-image $\Theta_t$ is successfully updated (Section~\ref{sec:preimage_update}). 
Passing these two tests, the target is considered visible with the pre-image accurately fitted. A filter is added if the object view has changed substantially and results in a new appearance (viewpoint). 
The change is measured by a difference between the reference aspect $\rho_0$ (i.e., a bounding box width-to-height ratio) and the aspect $\rho_t$ obtained from the current 2D projection of the object pre-image. 
Whenever this difference exceeds a threshold, i.e., $\| \rho_0 - \rho_t \| > \tau_\rho$, a new snapshot is created and the current ratio becomes a new reference, i.e., $\rho_0 \leftarrow \rho_t$.
Examples of images used to create separate DCF views are shown in Figure~\ref{fig:snapshots}.

\subsection{A multi-view DCF target detection} \label{sec:target_redetection}

Target presence is determined at each frame using the test described in Section~\ref{sec:target_presence_test}. Whenever the target is lost, the following re-detection mechanism is activated. At each frame all filters in the snapshot set $\{ \mathbf{h}^{(s)} \}_{s=1}^{S}$ are correlated with features extracted from a region centered at the last confident target position. To encode a motion model, the search region size is gradually increased in subsequent frames by a factor $\alpha^{\Delta t}_s$, where $\alpha_s >1$ is a fixed scale factor and $\Delta t$ is the number of frames since the last confident target position estimation. The correlation is efficiently calculated by padding the snapshots with zeros to the current search region size and applying FFT~\cite{lukezicFCLT}.
 
Since the target may change the size, a two-stage approach for re-detection is applied. 
First, the hypothesized target position is estimated as the location of the maximum correlation response and the filter $\mathbf{h}^{(m)}$ that yielded this response is identified. 
The current object scale is then computed as the ratio $s_f = \frac{D_{0}}{D_{t}}$ between the depth of the target in the first frame ($D_{0}$), and the depth $D_{t}$ at the current position. 
The depth is calculated  by the median of the D channel within the target bounding box. The filter that yielded the best correlation response ($\mathbf{h}^{(m)}$) is correlated again on the search region scaled by $s_f$ and target presence test is carried out (Section~\ref{sec:target_presence_test}). 
In case the test determines the target is present, the current filter is replaced, i.e., $\mathbf{h}_t \leftarrow \mathbf{h}^{(m)}$, and the re-detection process is deactivated.
  
\subsubsection{Target presence test}\label{sec:target_presence_test}

Recently, a target presence test has been proposed for long-term discriminative correlation filters~\cite{lukezicFCLT}. The test is based on computing tracking uncertainty value as a ratio $q_t = \frac{R_t}{\overline{R}}$ between the maximum correlation response in the current frame ($R_t$) and a moving average of these values in the recent $N_q$ frames when the target was visible. The test considers target lost whenever the ratio exceeds a pre-defined threshold $q_t > \tau_q$. It was showed in \cite{lukezicFCLT} that the test is robust to a range of thresholds.
  
To allow early occlusion detection, however,~\cite{Kart_ECCVW} introduce a test that compares the area of the segmentation mask with the area of the axis-aligned bounding box of the DCF. This test improves performance during occlusion, but gradual errors in scale estimation result in disagreement between the bounding box and the actual object and might lead to a reduced accuracy of the test.

The two tests are complementary and computationally very efficient, and the target presence is reported only if the considered target position passes the both tests.

\subsection{Object tracking by reconstruction}  \label{sec:tracker-summary}

Our object tracking by reconstruction approach (OTR) is summarized as follows.
 
{\bf Initialization.} 
The tracker is initialized from a bounding box in the first frame. Color and depth histograms are sampled as in~\cite{Kart_ECCVW} and a segmentation mask $\mathbf{m}$ is generated. 
The segmentation mask $\mathbf{m}$ is used to learn the initial filter $\mathbf{h}_0$ according to (\ref{eq:cf_cost_f_indep}), as well as to identify target pixels in the RGB-D model to initialize the pre-image $\Theta_0$ by~\cite{ruenz2017icra}. 
The set of snapshots is set to an empty set.

{\bf Localization.} 
A tracking iteration at frame $t$ starts with the target position $\mathbf{x}_{t-1}$ from the previous frame. 
A region is extracted around $\mathbf{x}_{t-1}$ in the current image and the position $\mathbf{x}_{t}$ with maximum correlation response is computed using the current filter $\mathbf{h}_{t-1}$ (along with all snapshots every $N_\mathrm{R}$ frames) as described in Section~\ref{sec:mvdcf}. 
The position $\mathbf{x}_{t}$ is tested using the target presence test from Section~\ref{sec:target_presence_test}. 
If the test is passed, the target is considered as well localized, and the visual models (i.e., filters and pre-image) are updated. 
Otherwise, target re-detection (Section~\ref{sec:target_redetection}) is activated in the next frame.

{\bf Update.} A color+depth segmentation mask $\mathbf{m}$ is computed within a region centered at $\mathbf{x}_{t}$ according to~\cite{Kart_ECCVW} to identify target pixels. The corresponding RGB-D pixels are used to update the pre-image $\Theta_t$, i.e.,~the 3D surfel representation along with its 3D pose (Section~\ref{sec:preimage_update}). 

The filter $\mathbf{h}_{t-1}$ is updated (\ref{eq:filter-update}) by the filter learned at the current position (\ref{eq:cf_cost_f_indep}) with support constraint computed from the pre-image (Section~\ref{sec:filter_constraint}). Finally, the target aspect change is computed using the updated pre-image and the set of snapshots are updated if significant appearance change is detected (Section~\ref{sec:mv_snapshots})

% ------------------------------------------------------------------------ %

\begin{table*}[!t]
\begin{center}
  \caption{Experiments on the Princeton Tracking Benchmark using the PTB protocol. Numbers in the parenthesis are the ranks.}
\label{table:resultsPTB} \scalebox{.69}
{
{\begin{tabular}{lrllllllllllllll}
%\toprule & \multicolumn{11}{c}{{\bf Attributes}}\\
\toprule
\specialcell{\bf Method} & \specialcell{\em Avg.\\ Rank} & \specialcell{\em Avg.\\ Success} & \specialcell{\em Human} & \specialcell{\em Animal} & \specialcell{\em Rigid} & \specialcell{\em Large} & \specialcell{\em Small} & \specialcell{\em Slow} & \specialcell{\em Fast} & \specialcell{\em Occ.} & \specialcell{\em No-Occ.} & \specialcell{\em Passive} & \specialcell{\em Active}\\
\midrule
\textit{OTR} & \bf{2.36} & \bf{0.769(1)} & 0.77(2) & 0.68(6) & 0.81(2) & 0.76(4) & \bf{0.77(1)} & 0.81(2) & \bf{0.75(1)} & 0.71(3) & 0.85(2) & \bf{0.85(1)} & 0.74(2)\\
%\textit{OTR} & \bf{2.36} & \bf{0.769(1)} & 0.77(2) & 0.68(6) & 0.81(2) & 0.76(4) & \bf{0.77(1)} & 0.81(2) & \bf{0.75(1)} & 0.71(3) & 0.85(2) & \bf{0.85(1)} & 0.74(2)\\
\textit{ca3dms+toh}~\cite{ca3dms} & 4.55 & 0.737(5) & 0.66(9) & 0.74(2) & \bf{0.82(1)} & 0.73(7) & 0.74(2) & 0.80(4) & 0.71(7) & 0.63(9) & \bf{0.88(1)} & 0.83(2) & 0.70(6)\\

\textit{CSR-rgbd++}~\cite{Kart_ECCVW} & 5.00 & 0.740(3) & 0.77(3) & 0.65(8) & 0.76(7) & 0.75(5) & 0.73(3) & 0.80(3) & 0.72(4) & 0.70(4) & 0.79(8) & 0.79(6) & 0.72(4)\\

\textit{3D-T}~\cite{Bibi3D} & 5.64 & 0.750(2) & \bf{0.81(1)} & 0.64(9) & 0.73(12) & \bf{0.80(1)} & 0.71(6) & 0.75(9) & 0.75(2) & \bf{0.73(1)} & 0.78(11) & 0.79(7) & 0.73(3)\\

\textit{PT}~\cite{princetonrgbd} & 6.09 & 0.733(6) & 0.74(6) & 0.63(11) & 0.78(3) & 0.78(3) & 0.70(7) & 0.76(5) & 0.72(6) & 0.72(2) & 0.75(13) & 0.82(4) & 0.70(7)\\

\textit{OAPF}~\cite{MESHGI_OAPF} & 6.09 & 0.731(7) & 0.64(12) & \bf{0.85(1)} & 0.77(6) & 0.73(8) & 0.73(5) & \bf{0.85(1)} & 0.68(9) & 0.64(8) & 0.85(3) & 0.78(9) & 0.71(5)\\

\textit{DLST}~\cite{DLST} & 6.45 & 0.740(4) & 0.77(4) & 0.69(5) & 0.73(13) & 0.80(2) & 0.70(9) & 0.73(11) & 0.74(3) & 0.66(6) & 0.85(4) & 0.72(13) & \bf{0.75(1)}\\

\textit{DM-DCF}~\cite{DMDCF} & 6.91 & 0.726(8) & 0.76(5) & 0.58(13) & 0.77(5) & 0.72(9) & 0.73(4) & 0.75(8) & 0.72(5) & 0.69(5) & 0.78(10) & 0.82(3) & 0.69(9)\\

\textit{DS-KCF-Shape}~\cite{Hannuna2016} & 7.27 & 0.719(9) & 0.71(7) & 0.71(4) & 0.74(9) & 0.74(6) & 0.70(8) & 0.76(6) & 0.70(8) & 0.65(7) & 0.81(6) & 0.77(11) & 0.70(8)\\

\textit{DS-KCF}~\cite{dskcf_bmvc} & 9.91 & 0.693(11) & 0.67(8) & 0.61(12) & 0.76(8) & 0.69(10) & 0.70(10) & 0.75(10) & 0.67(11) & 0.63(10) & 0.78(12) & 0.79(8) & 0.66(10)\\

\textit{DS-KCF-CPP}~\cite{Hannuna2016} & 10.09 & 0.681(12) & 0.65(10) & 0.64(10) & 0.74(10) & 0.66(12) & 0.69(12) & 0.76(7) & 0.65(12) & 0.60(12) & 0.79(9) & 0.80(5) & 0.64(12)\\

\textit{hiob-lc2}~\cite{hiob} & 10.18 & 0.662(13) & 0.53(13) & 0.72(3) & 0.78(4) & 0.61(13) & 0.70(11) & 0.72(12) & 0.64(13) & 0.53(13) & 0.85(5) & 0.77(12) & 0.62(13)\\

\textit{STC}~\cite{STC} & 10.45 & 0.698(10) & 0.65(11) & 0.67(7) & 0.74(11) & 0.68(11) & 0.69(13) & 0.72(13) & 0.68(10) & 0.61(11) & 0.80(7) & 0.78(10) & 0.66(11)\\
\bottomrule
\end{tabular}}}
\end{center}
\end{table*}

\section{Experimental analysis}\label{sec:experiments}

In this section, we validate OTR by a comprehensive experimental evaluation. The implementation details are provided in Section~\ref{sec:implementation}. 
Performance analysis on two challenging RGB-D datasets, Princeton Tracking Benchmark (PTB) and STC, is reported in Section~\ref{sec:resultsPTB} and Section~\ref{sec:resultsSTC}, respectively. 
Ablation studies are presented in Section~\ref{sec:ablation} to verify our design choices.

\subsection{Implementation details}\label{sec:implementation}

We use HOG features~\cite{HOG} and colornames~\cite{vandeWeijer2009} in our tracker. The parameters related to the tracker are taken from~\cite{Kart_ECCVW}. 
The ICP error threshold is empirically set to  $\tau_\mathrm{ICP}=5\cdot 10^{-4}$ and the aspect ratio change threshold is set to $\tau_\rho=0.20$. 
Maximum filter evaluation period is equal to $N_\mathrm{R}=5$ frames and $\alpha_s = 1.07$. 
All experiments are run on a single laptop with Intel Core i7 3.6GHz and the parameters for both tracking and 3D reconstruction are kept constant throughout the experiments. Our non-optimized implementation runs at $2$ fps. %however, we believe that this can be improved easily by using separate machines for pre-image $\Theta$ reconstruction and tracking. The proposed method runs its tracking module and the pre-image reconstruction module as separate instances on the same machine. The communication between these two are achieved by adopting TCP/IP protocol which allows extension for multiple machines straightforward.

\subsection{Performance on PTB benchmark~\cite{princetonrgbd} }\label{sec:resultsPTB}

The Princeton Tracking Benchmark~\cite{princetonrgbd} is the most comprehensive and challenging RGB-D tracking benchmark to date. The authors have recorded and manually annotated 100 RGB-D videos in real-life conditions using a Kinect v1.0. 
Ground truth bounding boxes of five sequences are publicly available whereas the ground truth for the remaining 95 sequences are kept hidden to prevent overfitting. 
Tracking performance is evaluated on the 95 sequences with the hidden ground-truth.
The sequences are grouped into 11 categories: \textit{Human}, \textit{Animal}, \textit{Rigid}, \textit{Large}, \textit{Small}, \textit{Slow}, \textit{Fast}, \textit{Occlusion}, \textit{No Occlusion}, \textit{Passive} and \textit{Active}. 
%The trackers are ordered by the rank averaged over the category ranks.
%The tracking output (axis-aligned bounding boxes) of 95 sequences is submitted to the evaluation server on the benchmark website to obtain the overall accuracy, accuracy per category and the overall ranking.

The performance is measured by employing a PASCAL VOC~\cite{PascalVOC2010} type of evaluation. Per-frame overlap $o_t$ is defined as
\begin{equation}\footnotesize
  o_t = \begin{cases}
    \frac{area(B_{TR} \cap  B_{GT})}{area(B {TR} \cup  B_{GT})}, \;\; \text{if both}  \; B_{TR} \; \text{and} \; B_{GT} \; \text{exist}\\
    1, \;\; \text{if neither} \; B_{TR} \; \text{and} \; B_{GT} \; \text{exists}\\
    0, \;\; \text{otherwise}
    \end{cases}
  \label{eq:pascalMeasure}
\end{equation}
where $B_{TR}$ is the output bounding box of the tracker and $B_{GT}$ is the ground truth bounding box.
Tracking performance is given as {\it success rate} which represents average overlap~\cite{cehovin_tip2016}. The PTB evaluation protocol sorts the trackers according to the primary performance measures with respect to each object category and computes the final ranking as the average over these ranks. In addition, the overall success rate is reported for detailed analysis.

The OTR tracker is compared to all trackers available on the PTB leaderboard: ca3dms+toh~\cite{ca3dms}, CSR-rgbd++~\cite{Kart_ECCVW}, 3D-T~\cite{Bibi3D}, PT~\cite{princetonrgbd}, OAPF~\cite{MESHGI_OAPF}, DM-DCF~\cite{DMDCF}, DS-KCF-Shape~\cite{Hannuna2016}, DS-KCF~\cite{dskcf_bmvc}, DS-KCF-CPP~\cite{Hannuna2016}, hiob\_lc2~\cite{hiob} and we added two recent trackers STC~\cite{STC} and DLST~\cite{DLST}. 
Results are reported in Table~\ref{table:resultsPTB}. 
 
OTR convincingly sets the new \textit{state-of-the-art} in terms of both overall ranking and the average success by a large margin compared to the next-best trackers (Table~\ref{table:resultsPTB}). 
In terms of average success, OTR obtains a $4.3\%$ gain compared to the second ranking tracker ca3dms+toh~\cite{ca3dms}, which tracks the target in 3D as well, but without reconstruction. 
This result speaks in favour of our 3D-based pre-image construction and its superiority for RGB-D tracking.

In addition to being the top overall tracker, the performance of OTR is consistent across all categories. 
OTR is consistently among the top trackers in each category and achieves the top rank in three categories and the second best in five categories. 
This suggests that our tracker does not overfit to a certain type of scenario and it generalizes very well unlike some other methods in the benchmark.

A closely related work to our own is recent CSR-rgbd++, which combines a single CSR-DCF with color and depth segmentation and implements a target re-detection. 
OTR obtains a significant $6.6\%$ increase over CSR-rgbd++ in \textit{Rigid} category, which speaks in favor of our DCFs approach with several views connected to a 3D pre-image that localizes the target more precisely. 
%A detailed analysis of the proposed set of view-specific DCFs shows that on average $3.9$ views were generated per tracking sequence. 
On the \textit{No-Occ.} category, OTR outperforms CSR-rgbd++ by a $7.6\%$ success rate. This can be attributed to the advantage of using a pre-image $\Theta$ for DCF training described in Section~\ref{sec:filter_constraint}. 
 
%  The average numbers of DCFs $7.9$ and $1.4$ for \textit{Rigid} and \textit{Deformable} objects respectively 

%An example showing the power of multi-view DCF formulation is given in Fig.~\ref{fig:bear_mvdcf}. On the top row, the localization process fails to accurately locate the target after *****K***** frames and a full 360 degrees rotation without the multi-view DCF. This is due to the fact that the appearance model that is seen initially is averaged and thus, the information about those frames is lost. On the bottom row, thanks to the fact that the system has stored the previously seen view, multi-view DCF is able to locate the target perfectly. .

%\begin{figure}[h]
%  \begin{center}
%    \includegraphics[width=1.0\linewidth]{resources/bear_mvdcf.png}
%    \caption{MV-DCF is better than DCF}
%      \label{fig:bear_mvdcf}
%  \end{center}
%\end{figure}

%\begin{figure}[h]
%  \begin{center}
%    \includegraphics[width=1.0\linewidth]{latex/resources/ptb_rigid_filters.png}
%    \includegraphics[width=1.0\linewidth]{latex/resources/ptb_deformable_filters}
%    \caption{Number of filters created in PTB dataset. \cmnt{[AL] I think that giving only statistics about this (mean %and potentially std. deviation) is enough. Graphs do not tell much, they might just encourage tricky questions of the reviewers.}}
%      \label{fig:bear_mvdcf}
%  \end{center}
%\end{figure}

\begin{table*}[ht]
\begin{center}
  \caption{The normalized area under the curve (AUC) scores computed from one-pass evaluation on the STC Benchmark~\cite{STC}.}
\label{table:resultsSTC}\scalebox{0.9}
{
{\begin{tabular}{llllllllllll}
\toprule \multicolumn{12}{c}{\bf Attributes}\\
{\bf Method} & {\em AUC} & {\em IV} & {\em DV} & {\em SV} & {\em CDV} & {\em DDV} & {\em SDC} & {\em SCC} & {\em BCC} & {\em BSC} & {\em PO}\\
\midrule
\textit{OTR} & \bf{0.49} & \bf{0.39} & \bf{0.48} & \bf{0.31} & 0.19 & \bf{0.45} & \bf{0.44} & 0.46 & \bf{0.42} & \bf{0.42} & \bf{0.50}\\
\textit{CSR-rgbd++}~\cite{Kart_ECCVW} & 0.45 & 0.35 & 0.43 & 0.30 & 0.14 & 0.39 & 0.40 & 0.43 & 0.38 & 0.40 & 0.46\\
\textit{ca3dms+toh}~\cite{ca3dms} & 0.43 & 0.25 & 0.39 & 0.29 & 0.17 & 0.33 & 0.41 & \bf{0.48} & 0.35 & 0.39 & 0.44\\
\textit{STC}~\cite{STC} & 0.40 & 0.28 & 0.36 & 0.24 & \bf{0.24} & 0.36 & 0.38 & 0.45 & 0.32 & 0.34 & 0.37\\
\textit{DS-KCF-Shape}~\cite{Hannuna2016} & 0.39 & 0.29 & 0.38 & 0.21 & 0.04 & 0.25 & 0.38 & 0.47 & 0.27 & 0.31 & 0.37\\
\textit{PT}~\cite{princetonrgbd} & 0.35 & 0.20 & 0.32 & 0.13 & 0.02 & 0.17 & 0.32 & 0.39 & 0.27 & 0.27 & 0.30\\
\textit{DS-KCF}~\cite{dskcf_bmvc} & 0.34 & 0.26 & 0.34 & 0.16 & 0.07 & 0.20 & 0.38 & 0.39 & 0.23 & 0.25 & 0.29\\
\textit{OAPF}~\cite{MESHGI_OAPF} & 0.26 & 0.15 & 0.21 & 0.15 & 0.15 & 0.18 & 0.24 & 0.29 & 0.18 & 0.23 & 0.28\\
\bottomrule
\end{tabular}}}
\end{center}
\end{table*}

\subsection{Performance on STC benchmark~\cite{STC}}\label{sec:resultsSTC}

The STC benchmark~\cite{STC} has been recently published to complement the PTB benchmark in the number of categories and diversity of sequences.~The sequences are recorded using Asus Xtion sensors and the authors annotated every frame of every video with $10$ attributes; \textit{ Illumination variation} (IV), \textit{Depth variation} (DV), \textit{Scale variation} (SV), \textit{Color distribution variation} (CDV), \textit{Depth distribution variation} (DDV), \textit{Surrounding depth clutter} (SDC), \textit{Surrounding color clutter} (SCC), \textit{Background color camouflages} (BCC), \textit{Background shape camouflages} (BSC), \textit{Partial occlusion} (PO). 
These attributes were either automatically computed or manually annotated.

The tracking performance is measured by precision and success plots computed from a one-pass evaluation akin to~\cite{OTB}. 
Success plot shows the portion of correctly tracked frames with respect to the different values of the overlap thresholds. Tracking performance is measured by a non-normalized area under the curve on this graph, i.e., the sum of values on the plot. The standard AUC measure~\cite{OTB} is obtained by dividing the non-normalized AUC by the number of overlap thresholds. 
The number of thresholds is the same for all evaluated trackers and only scales the non-normalized AUC to interval $[0,1]$. We therefore report the standard AUC values, which is the more familiar measure in the tracking community. 
Precision plot is constructed similarly to success plot, by measuring the portion of frames with center-error smaller than a threshold. The overall measure on precision plot is computed as the value at 20 pixels error threshold.

\begin{figure}[h]
  \begin{center}
    \includegraphics[width=0.8\linewidth]{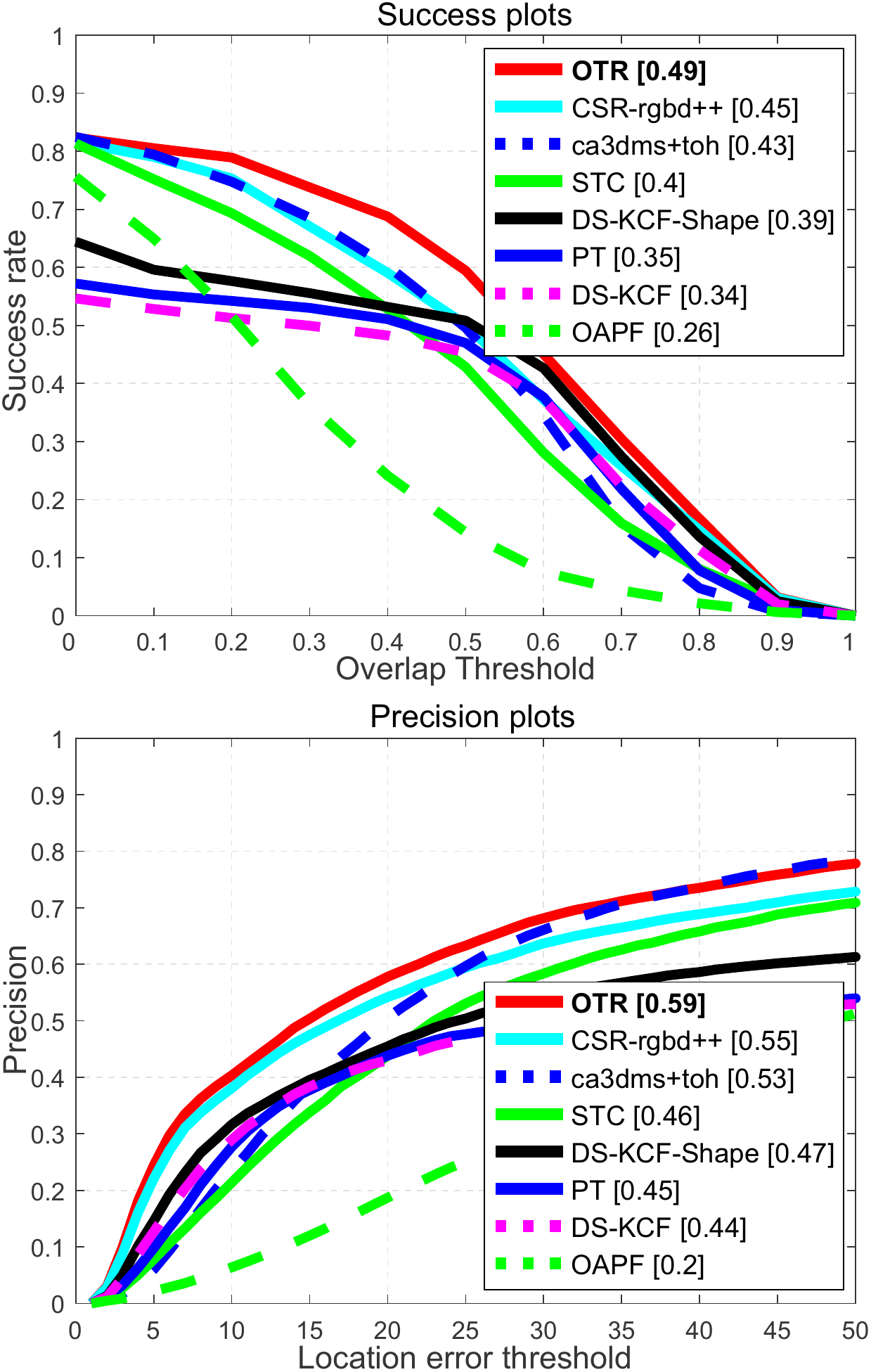}
    \caption{Success and precision plots on STC benchmark~\cite{STC}.}
      \label{fig:STC_AUC_Precision}
  \end{center}
\end{figure}

\begin{figure}[!t]
  \begin{center}
    \includegraphics[width=\linewidth]{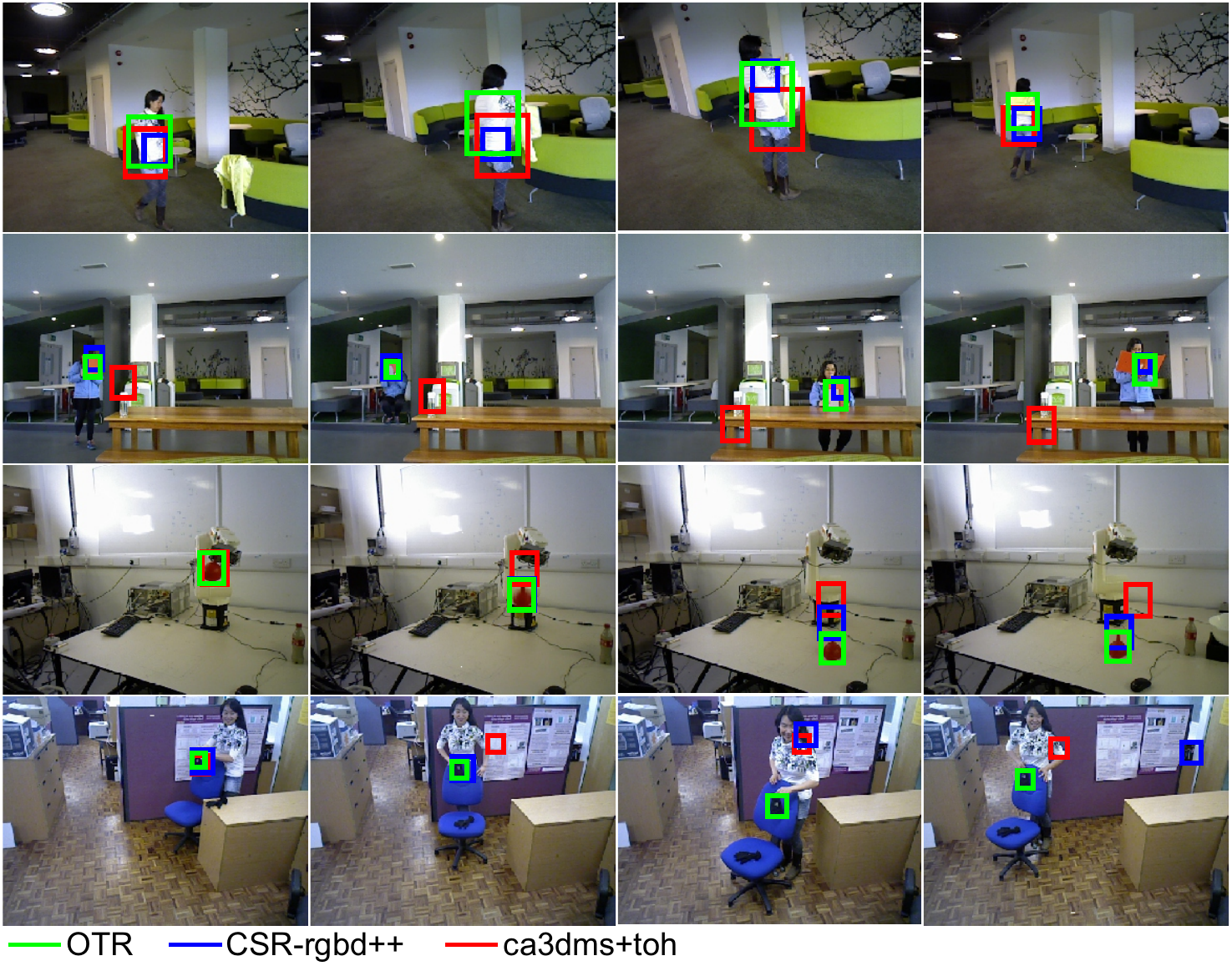}
    \caption{Tracking results on four sequences from STC dataset~\cite{STC}. The proposed OTR tracker confidently tracks the target undergoing a substantial pose change. Two state-of-the-art RGB-D trackers (CSR-rgbd++~\cite{Kart_ECCVW} and cs3dms+toh~\cite{ca3dms}), that do not apply the multi-view DCFs nor target 3D pre-image, result in less accurate localization or failure.\vspace{-0.5cm} 
}
      \label{fig:tracking-examples}
  \end{center}
\end{figure}

The OTR tracker is compared to the following trackers: CSR-rgbd++~\cite{Kart_ECCVW}, ca3dms+toh~\cite{ca3dms}, STC~\cite{STC}, DS-KCF-Shape~\cite{Hannuna2016}, PT~\cite{princetonrgbd}, DS-KCF~\cite{dskcf_bmvc} and OAPF~\cite{MESHGI_OAPF}.
The results are presented in Table~\ref{table:resultsSTC} and Figure~\ref{fig:STC_AUC_Precision}.
As on PTB benchmark (Section~\ref{sec:resultsPTB}), OTR outperforms the \textit{state-of-the-art} by a large margin not only in the overall score but in most of the categories except \textit{CDV} \textit{(Color Distribution Variation)} and \textit{SCC} \textit{(Surrounding Color Clutter)}, where it is ranked among top three trackers. 
The overall top performance and excellent per-attribute performance support our observations on PTB benchmark that OTR is capable of handling various tracking scenarios and generalizes well over the different datasets. 
Qualitative tracking results on the four sequences from STC dataset are shown in Figure~\ref{fig:tracking-examples}.
%This strongly supports our earlier observation of generalization. 
%The average number of views created per sequence on this dataset is $3.08$.

%Our results on STC Dataset are given in Table~\ref{table:resultsSTC} and Fig.~\ref{fig:STC_AUC_Precision}. The numbers of CSR-rgbd++~\cite{Kart_ECCVW}, STC~\cite{STC}, DS-KCF-Shape~\cite{Hannuna2016}, PT~\cite{princetonrgbd}, DS-KCF~\cite{dskcf_bmvc} and OAPF~\cite{MESHGI_OAPF} are taken from the benchmark paper~\cite{STC}. ca3dms+toh~\cite{ca3dms} and CSR-rgbd++~\cite{Kart_ECCVW} are run by us using the source code provided by the authors. As on PTB, the proposed method outperforms the \textit{state-of-the-art} by a large margin not only in overall results but every category except \textit{CDV} \textit{(Color Distribution Variation)} and \textit{SCC} \textit{(Surrounding Color Clutter)}. This strongly supports our earlier observation of generalization. The average number of views created per sequence on this dataset is $3.08$.

%------------------------------------------------------------------------
\subsection{Ablation studies}\label{sec:ablation}

\begin{table*}[!h]
\begin{center}
  \caption{Ablation studies on the PTB benchmark~\cite{princetonrgbd}.}
\label{table:ablationPTB} \scalebox{0.90}
{
{\begin{tabular}{lrllllllllllllll}
\toprule 
%& \multicolumn{11}{c}{{\bf Attributes}}\\
{\bf Method} & \specialcell{\em Avg.\\Success} & {\em Human} & {\em Animal} & {\em Rigid} & {\em Large} & {\em Small} & {\em Slow} & {\em Fast} & {\em Occ.} & {\em No-Occ.} & {\em Passive} & {\em Active}\\
\midrule
\textit{OTR} & \bf{0.769} & 0.77 & \bf{0.68} & \bf{0.81} & \bf{0.76} & \bf{0.77} & \bf{0.81} & \bf{0.75} & \bf{0.71} & \bf{0.85} & \bf{0.85} & \bf{0.74}\\
\textit{OTR$_\mathrm{-VS}$} & 0.743 & 0.75 & 0.66 & 0.77 & 0.74 & 0.74 & 0.79 & 0.72 & 0.67 & 0.84 & 0.81 & 0.72\\
\textit{OTR$_\mathrm{-3D-VS}$} & 0.740 & \bf{0.78} & 0.61 & 0.76 & 0.75 & 0.73 & 0.79 & 0.72 & \bf{0.71} & 0.78 & 0.79 & 0.72\\
%\textit{OTR-DCF} & 0.265 & 0.75 & 0.65 & 0.76 & 0.74 & 0.73 & 0.79 & 0.71 & 0.68 & 0.82 & 0.81 & 0.71\\
\bottomrule
\end{tabular}}}
\end{center}
\end{table*}

The main components of our tracker are (i) the 3D-based pre-image, which provides an improved target segmentation, (ii) the set of multiple view-specific target DCFs and (iii) the interaction between the former two components. An ablation study is conducted on the PTB~\cite{princetonrgbd} dataset to evaluate the extent of contribution of each component.
We implemented two variants of the proposed tracker with the 3D pre-image and view-specific DCFs, denoted as OTR. The first variant is the tracker without the view-specific DCFs, denoted as ~OTR$_\mathrm{-VS}$. The second variant is the tracker without the view-specific DCFs and without the 3D pre-image OTR$_\mathrm{-3D-VS}$.

%A detailed analysis of the proposed set of view-specific DCFs shows that on average $3.9$ views were generated per tracking sequence. 

The results of the ablation study are reported in Table~\ref{table:ablationPTB}. The proposed OTR with all components achieves a $0.769$ success rate. 
Removing the view-specific target representation (OTR$_\mathrm{-VS}$) results in nearly $3\%$ success rate drop in tracking performance ($0.743$ success rate). 
Removing both view-specific and 3D pre-image representation (OTR$_\mathrm{-3D-VS}$) further reduces the tracking performance to $0.740$ success rate. 

On the \textit{Occlusion} category the OTR tracker outperforms the version without a view-specific formulation (OTR$_\mathrm{-VS}$) by $6\%$ increase in the success rate. The view-specific set of DCFs \textit{remembers} the target appearance from different views, which helps in reducing drifting and improves re-detection accuracy after occlusion. On average, $4$ views were automatically generated by the view-specific DCF in OTR per tracking sequence. 
The tracker version without the view-specific formulation {\it forgets} the past appearance, which reduces the re-detection capability.

%In addition to that we measured the average number of alternative views created in each sequence. There were average $3.9$ and $3.08$ views created on PTB~\cite{princetonrgbd} and STC~\cite{STC}, respectively. 

%This quantitatively proves that multi-view DCF formulation has an essential impact on the tracking performance which is an expected result because multiple appearances are checked and the most similar is used for tracking. 
%This is especially useful when target appearance changes during occlusion and it looks similar than in the past (before occlusion). 

In situations without occlusion, the 3D pre-image plays a more important role than the view-specific DCF formulation. 
Removing the 3D pre-image creation from the tracker results in 7\% success rate reduction, which indicates the significant importance of using the 3D pre-image for robust DCF learning. 

Overall, the addition of 3D pre-image and view-specific target representation improves performance of the baseline version OTR$_\mathrm{-3D-VS}$ by approximately $4\%$ in tracking success rate.
The ablation study results conclusively show that every component importantly contributes to the tracking performance boost.

\section{Conclusions}\label{sec:conclusion}

A new long-term RGB-D tracker, called OTR --  Object Tracking by Reconstruction is presented. The target 3D model, a pre-image, is constructed by a surfel-based ICP.
The limited convergence range of the ICP and the requirement to automatically identify object pixels used for reconstruction is addressed by utilizing a DCF for displacement estimation and for approximate target segmentation. The 3D pre-image in turn constrains the DCF learning, and is used for generating view-specific DCFs. These are used for localization as well as for target re-detection, giving the tracker a long-term tracking quality.

The OTR tracker is extensively evaluated on two challenging RGB-D tracking benchmarks and compared to 12 \textit{state-of-the-art} RGB-D trackers.~OTR outperforms all trackers by a large margin, setting a new \textit{state-of-the-art} on these benchmarks. 
An ablation study verifies that the performance improvements come from the 3D pre-image construction, the view-specific DCF set and the interaction between the two.

The view-specific DCF formulation allows long-term tracking of poorly textured and small objects over large displacements. Our future work will focus on extension to model-based tracking with pre-learned models on realistic, open-world scenarios. In addition, we plan to consider improvements by ICP robustification and deep features.
 
%We introduced object tracking by reconstruction with view-specific discriminative correlation filters. To the best of our knowledge, the proposed approach is the first to introduce a joint framework for object pre-image estimation and usage of it in DCF model learning. 
%Our method exploits the 3D structure of the target object twofolds. Firstly, a novel mask constraint (spatial support) for DCF training is proposed using a 3D pre-image formulation. This provides a more precise definition of the target object compared to the segmentation based approaches. The second novelty we propose is adopting 3D pre-image for detecting 3D pose changes to overcome the problems in traditional object trackers. This allows us to distinguish rotations from occlusions. Furthermore, we use the rotation information to store appearances of the target under different 3D poses via a view-specific DCF formulation. As a result, contrary to the existing trackers in the literature, the proposed method does not \textit{forget} the earlier appearances of the target object. Experiments on PTB~\cite{princetonrgbd} and STC~\cite{STC} datasets showed that our formulation provided significant improvements over the \textit{state-of-the-art} on both.

%------------------------------------------------------------------------
{\small
\bibliographystyle{ieee}
\bibliography{egbib}
}

\end{document}